\documentclass[10pt,twocolumn,letterpaper]{article}

\usepackage{cvpr}              
\usepackage{booktabs}
\usepackage{multirow}
\usepackage{adjustbox}
\usepackage{subcaption}
\usepackage{comment}
\usepackage{amssymb}
\usepackage{mathrsfs}
\usepackage{mathtools}
\usepackage[accsupp]{axessibility}  
\usepackage{subcaption}
\usepackage{smartdiagram}
\usesmartdiagramlibrary{additions}

\definecolor{cvprblue}{rgb}{0.21,0.49,0.74}
\usepackage[pagebackref,breaklinks,colorlinks,allcolors=cvprblue]{hyperref}
\title{Ghosts in the Point Clouds: De-glaring LiDAR in the Transient Domain}
\author{
Avery Gump* \quad
Connor Henley* \quad
Sungjin Cheong \quad
Akarsh Prabhakara \quad
Mohit Gupta\\
University of Wisconsin--Madison, Madison, WI, USA\\
{\tt\small \{gump, connor.henley, sungjin.cheong, aprabhakara, mgupta37\}@wisc.edu}
}

\begin{document}

\twocolumn[{
 \renewcommand\twocolumn[1][]{#1}
\maketitle
 \vspace{-1cm}
 \begin{center}
 \begingroup
 \hypersetup{
     colorlinks=true,
     urlcolor=[rgb]{1,0.2,0.6},
 }
 \href{https://wisionlab.com/project/deglaring-LiDAR}{%
 \setlength{\fboxsep}{1.5pt}
   \color{white}\fbox{\strut\textcolor[rgb]{1,0.2,0.6}{wisionlab.com/project/deglaring-LiDAR}}%
 }
 \endgroup
 \end{center}
 }]

\begin{abstract}
\par\vskip -1em
Modern LiDARs are rapidly transitioning from bulky, mechanically scanned systems to ultra-compact, low-cost, solid-state arrays. This miniaturization—while enabling scalability, affordability, and camera-like data structures—introduces a new and severe failure mode: internal-multipath glare. When light from a bright or retroreflective surface reflects and scatters within the LiDAR, light that should reach a single pixel spreads across the pixel array. The resulting artifacts create phantom objects, obscure real ones, and produce safety-critical ``ghosts in the point clouds.'' This paper introduces a physically grounded sensing model and algorithmic techniques for addressing this effect. We show that internal glare can be represented as a linear, scene-independent operator—the Transient Glare Spread Function (TGSF)—acting on the transient measurements. Building on this model, we develop a training-free approach that operates on low-level LiDAR detections (or echoes) prior to point-cloud formation, leveraging knowledge of the glare spread function to reason about the likelihood of each detection arising from glare. The resulting approach is compatible with existing LiDAR signal-processing pipelines, and deployable on unmodified commercial sensors. Using experiments with real single-photon LiDAR hardware, we demonstrate substantial suppression of severe glare artifacts while preserving true scene structure.
\begingroup
\renewcommand*{\thefootnote}{\fnsymbol{footnote}}
\footnotetext[1]{Equal contribution}
\footnotetext[2]{This research was supported by NSF CAREER Award 1943149, NSF grant CNS-2107060, NRT Award Number 2152163, and ONR grant N000142412155. We thank Sacha Jungerman for helpful discussions.}

\endgroup

\end{abstract}    
\vspace{-.5cm} 
\section{A Changing LiDAR Landscape}\label{sec:intro}
\vspace{-.15cm}
LiDAR technology is undergoing a rapid and fundamental transformation. The field is moving away from legacy scanning-based systems—bulky, expensive, and mechanically complex—toward compact, solid-state architectures that are low-cost and scalable. This transition, driven in large part by the emergence of high-resolution \emph{Single-Photon Avalanche Diode (SPAD)} arrays~\cite{Charbon:2013:SPAD}, promises to transform LiDAR from a “necessary evil” into a reliable and inexpensive modality, much like “just another, but more informative, camera.” Furthermore, dense pixel arrays in future LiDARs will make their data resemble camera imagery in structure and scale, potentially merging camera and LiDAR processing pipelines. 

\begin{figure}[t]
\centering
\vspace*{0.1cm}
\includegraphics[width=\columnwidth]{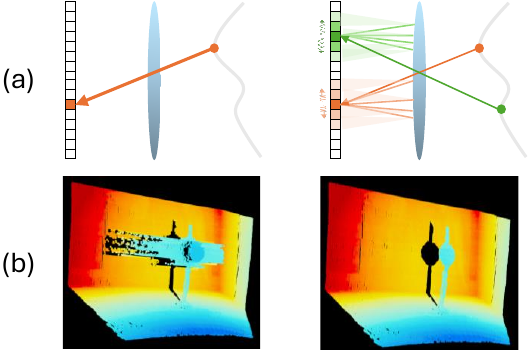}
\vspace*{-0.3cm}
\caption{(a) For an ideal LiDAR sensor, there is a one-to-one correspondence between scene point and sensor pixel. (b) ``Camera-fication'' of LiDAR leads to glare from internal optical scattering, multiple internal reflections, and electrical crosstalk.
}
\vspace*{-0.4cm}
\label{fig:glare_concept}
\end{figure}

\begin{figure*}[t]
\centering
\vspace*{0.1cm}
\includegraphics[width=0.97\textwidth]{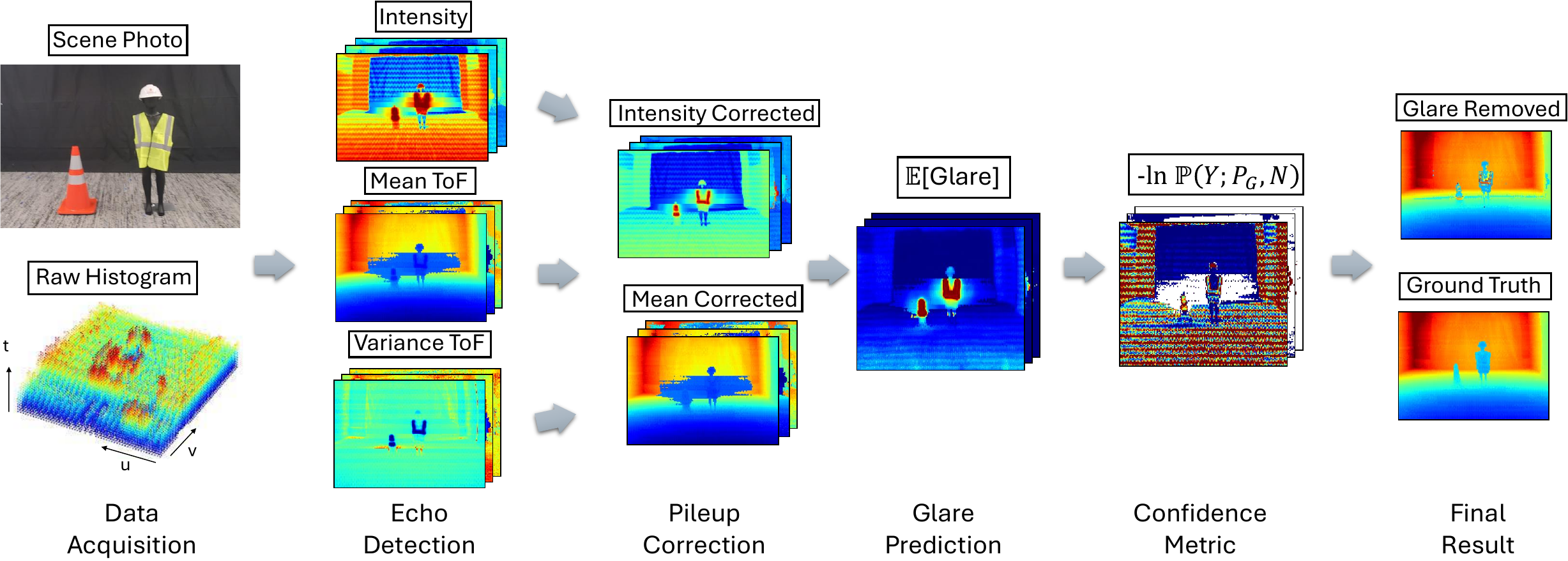}
\vspace*{-0.3cm}
\caption{Overview of our glare mitigation method.  First, raw histograms are processed to extract intensity, mean time-of-flight (ToF), and ToF variance of several echoes per pixel.  Second, a pileup correction method corrects the intensity and ToF of the brightest returns, which induce glare.  Third, a glare model predicts the glare contribution to each echo.  Finally, a confidence metric quantifies the likelihood that each echo contains signal in excess of the predicted glare.  We choose the highest confidence echo at each pixel to form a depth map.}
\vspace*{-0.4cm}
\label{fig:teaser}
\end{figure*}


This \emph{camera-fication} of modern LiDARs comes with a cost. As in conventional cameras, modern LiDAR arrays view the world through shared optical elements. When a bright return enters the system, light that should ideally be focused onto a single pixel can undergo multiple internal reflections between the lens and sensor, and can scatter within the optical assembly itself, redistributing itself across a wider neighborhood of pixels. The result is an increasingly critical phenomenon we refer to as \textbf{internal-multipath glare} (also called \emph{blooming} in the LiDAR community).\footnote{Unlike glare in passive imaging~\cite{veiling_glare_in_hdr,astronomicalGlare}—typically driven by strong external sources such as sun or headlights in autonomous driving or bright stars in astronomy—LiDAR glare is self-induced by intense returns.} The redistribution creates false or “phantom” structures while suppressing true returns from real surfaces. This results in LiDARs \emph{seeing what is not there} (e.g., a phantom wall materializing on a freeway due to overhead signs that cause an autonomous vehicle to \emph{brake}), and simultaneously \emph{missing what is there} (e.g., a child standing beneath a retroreflective stop sign could be obscured). Such `ghosts in the point clouds’ are safety-critical failure modes. One of the most common culprits are retroreflectors (road signs, license plates, lane markers, safety vests, and construction cones). These objects are deliberate and ubiquitous components of modern infrastructure; constructing a realistic scenario \emph{without} glare sources may indeed be harder than constructing one with them.



Why has this failure mode remained largely invisible to the research community? Historically, legacy LiDARs avoided severe optical crosstalk by design. Commercial scanning systems such as the Velodyne HDL-64e grouped detectors behind separate optical paths and—critically—fired only one detector per group at any given time.\footnote{For example, in the Velodyne HDL-64e, two groups of 32 detectors each share separate optics, and only one detector in each group is active at a time~\cite{VelodyneManual}. This design choice was motivated explicitly by the need to prevent blooming when encountering retroreflectors.} This combination of dedicated optics and scanning mechanisms effectively suppressed internal reflections, but at significant cost: large form factors, spinning mechanisms with higher cost/complexity, sparse point clouds, motion-induced artifacts, and limited resolution. Indeed, the severity of blooming was such that prior LiDAR generations adopted fundamentally sub-optimal hardware architectures to avoid it. Consequently, widely used datasets and benchmarks~\cite{KITTI,nuscenes,waymo_open_dataset}, captured using such systems, exhibit minimal glare. The landscape is now changing. Modern solid-state LiDARs unlock high sensitivity, fine temporal resolution, and high-resolution arrays at low costs. However, these architectures alter internal light transport and acquisition, making internal glare not a corner case but a frequent and severe artifact.


Internal glare breaks a foundational assumption underlying LiDAR processing: the one-to-one correspondence between a pixel and a single scene point (\cref{fig:glare_concept}). Once light from a bright return spreads across pixels, downstream steps (peak-finding, denoising, and point-cloud processing) operate on corrupted data. Waiting to address glare until after point cloud formation is often too late: the artifacts, becoming entangled with geometry, are difficult to separate from true scene elements, and retroreflector‐driven ghosts can appear more confident than the real objects. \smallskip

We argue that internal glare should be addressed early in the sensing pipeline, \emph{before} point-cloud formation. Modern single-photon (SP) LiDARs~\cite{Charbon:2013:SPAD,lee2023caspi,Lindell} capture rich temporal information in the form of transient histograms. In this measurement space, the physics of internal glare can be modeled as a \emph{linear, scene-independent operator} acting on an ideal, glare-free transient histogram. We refer to this operator as the \emph{Transient Glare Spread Function (TGSF)}, which describes how photons from a single pixel are redistributed across other pixels and neighboring time bins through internal reflections and scattering.\footnote{In practice, the temporal extent of this spreading is typically small, and the effect can be treated as predominantly spatial; thus, we refer to the operator as the \emph{Glare Spread Function (GSF)} in the rest of the paper.}

Building on this observation, we develop LiDAR deglare algorithms that operate directly on low-level temporal peak detections (or \emph{echoes}) derived from transient histograms, prior to point-cloud formation (\cref{fig:teaser}). We leverage knowledge of the GSF as a forward model to reason about the likelihood that each detected echo arises from internal glare. This allows us to identify and suppress glare-induced detections while preserving true scene returns. Importantly, the method integrates naturally with standard LiDAR signal-processing pipelines and can be deployed on unmodified, commercially available sensors.

A key challenge in this regime is that strong glare—particularly from retroreflectors—is typically accompanied by severe nonlinear distortion due to photon pileup~\cite{coates_correction,pileup_anant,pileup_anant,Heide:2018:subpicosecond}. The proposed approach therefore jointly addresses both internal glare and pileup effects, and is guided by the following principles:

\begin{itemize}
    \item \textbf{Generality.} A single method applies across retroreflector types, orientations, sizes, counts, and scene configurations—without requiring explicit detection or modeling of glare sources.
    \item \textbf{Training-free.} The method does not require large-scale datasets, making it well-suited to single-photon LiDAR settings where labeled data is scarce.
    \item \textbf{Compatibility with downstream processing.} The method integrates seamlessly into existing LiDAR signal-processing pipelines prior to point-cloud formation, and can propagate confidence measures alongside each echo.
    \item \textbf{Efficiency.} The approach is computationally lightweight.
\end{itemize}

We demonstrate, using a commercial SP-LiDAR evaluation kit, that internal glare produces severe artifacts that standard pipelines do not remove—phantom returns and “ghost” objects that occlude real ones. Across scenes containing diverse and multiple retroreflectors, our method substantially suppresses these effects while preserving true structure, even in the presence of strong pileup distortion. \smallskip

\noindent{\bf Contributions.} This paper makes four key contributions:
\begin{itemize}
    \item Identification of \emph{internal-multipath glare} as a critical failure mode for next-generation solid-state LiDARs.
    \item A transient-domain formulation of glare that accounts for both internal multipath and nonlinear pileup distortion arising from high-intensity returns.
    \item A training-free algorithm that leverages knowledge of the glare spread function to estimate the likelihood that each LiDAR detection is induced by glare.
    \item A practical approach that operates prior to point-cloud formation, and integrates with standard LiDAR pipelines.
\end{itemize}

\vspace{-.25cm}
\section{Related Work} \label{sec:related}

\vspace{-.15cm}
\noindent \textbf{Glare in passive imaging and astronomy.}
In conventional cameras, glare (often termed \emph{veiling glare}) has been modeled and mitigated using linear optical spread models and inverse formulations. Talvala \etal~\cite{veiling_glare_in_hdr} measured a glare spread function and applied deconvolution for glare removal. Glare has also been studied in astronomical imaging, which faces extreme dynamic range~\cite{astronomicalGlare}. Optical filtering has been explored to improve the conditioning of the glare spread function~\cite{glare_encoding_hdr}. As in these works, we use a linear GSF to model glare formation, but also consider the transient nature of glare in LiDAR, as well as nonlinear pileup effects unique to SP-LiDAR. \smallskip

\noindent \textbf{Glare in LiDAR.} Work specifically addressing LiDAR internal glare is sparse. Recent learning-based approaches operate on point clouds to detect or correct blooming artifacts~\cite{BloomNet,deep_learning_bloom_correction}, typically relying on synthetic training data and strong assumptions about reflector geometry. Because these methods act \emph{after} point-cloud formation, artifacts are already entangled with geometry, and such approaches often struggle to generalize across reflector types, layouts, and intensities. By contrast, we use a glare model that depends only on internal sensor characteristics.  As such, our method is training-free and generalizes to arbitrary scenes.

Electrical crosstalk in SPAD arrays, where secondary photons generated during an avalanche re-trigger neighboring pixels~\cite{spad_crosstalk_optica,spad_crosstalk_trench,spad_crosstalk_jsap}, has also been studied. However, such electrical crosstalk is typically confined to a small neighborhood of adjacent pixels, and modern devices largely mitigate this effect using deep trench isolation. Internal glare on the other hand arises from optical reflections and scattering between the SPAD array and shared optics, and can spread over large portions of the sensor, producing spatially extended artifacts. \vspace{1pt}

\noindent \textbf{Computational single-photon imaging.}
A rapidly growing body of work addresses single-photon LiDAR artifacts orthogonal to glare: pileup modeling and mitigation~\cite{pileup_anant,Heide:2018:subpicosecond}, probabilistic inference and denoising for photon-limited histograms~\cite{Shin,Lindell,lee2023caspi,eccv_noise,bhavya_iccv,rapp_and_Goyal}, and compression/efficient streaming of transient data~\cite{felipe_compression,atul_compression1,atul_compression2,sheehan2024}. These advances are complementary to our goals.  Our method for identifying and suppressing glare-induced returns is compatible with existing pileup correction, denoising, peak-finding, compression, and downstream learning—suggesting a modular \emph{photon-processing pipeline} for next-generation LiDAR.
\vspace{-.25cm}
\section{Method}
\label{sec:method}
\vspace{-.25cm}
In Sec.~\ref{sec:transientglare} we propose a transient-domain glare model for glare in ``camera-fied'' LiDAR sensors.  In \ref{sec:pileup} we describe our method for correcting photon pileup, an artifact that plagues extremely high-flux signals measured by SP-lidars, and usually co-occurs with glare.  In \ref{sec:echoid}, we review our method for mitigating glare in two steps: first, by predicting glare contributions to each detected echo; second, by determining the likelihood that each echo contains a true (non-glare) signal.  Our full pipeline is visualized in Fig.~\ref{fig:teaser}.
\vspace{-.15cm}
\subsection{Glare in the Transient Domain}
\label{sec:transientglare}
\vspace{-.15cm}
Veiling glare in optical imaging systems arises from a variety of optical processes, including multiple reflections between lens and image sensor surfaces, and diffuse scattering from lens surfaces and aperture edges \cite{ISO9358}.  In photon-counting lidar systems, electroptical processes such as photon re-emission due to hot carriers relaxation can also produce a glare-like effect, often referred to as ``crosstalk'' \cite{spad_crosstalk_optica, spad_crosstalk_trench}.  Several sources of glare are depicted in Fig.~\ref{fig:glare_concept}a,b.  In spite of its myriad causes, glare formation is a linear process, in that the intensity of the glare signal responds linearly to the intensity of incident light.

We consider a camera-like, direct-time-of-flight (dToF) LiDAR sensor that consists of a time-resolving detector array positioned behind a lens in an imaging configuration, and a transmitter that illuminates the scene with pulsed laser light in an arbitrary spatial pattern (e.g. flash, line scanned, grid of dots, etc.).  The raw measurement captured by dToF cameras is a three-dimensional datacube that resolves the spatial (pixel) position of incident light $(u,v)$ as well as its time of flight $t$.  This means that, unlike in regular photography, glare observed by a dToF camera has a transient aspect---the time that glare is observed depends on the depth of the bright surface that produces it.

We define $y$ as the transient intensity of light incident upon the image sensor that has been corrupted by glare and noise, and $x$ as the ``true'' uncorrupted transient intensity.  The process of glare formation is governed by the six-dimensional \emph{transient glare spread function} (TGSF) $a(u,v,t,u',v',t')$, which quantifies the linear response of the measured datacube $y$ at coordinate $(u,v,t)$ to the uncorrupted intensity of $x$ at $(u',v',t')$.  Altogether the measurement process can be described by the following integral equation
\vspace{-.25cm}
\begin{multline}
    y(u,v,t) = \\ \iiint a(u,v,t,u',v',t')x(u',v',t')du'dv'dt' + \eta(u,v,t) .
    \label{eq:st_integral}
\end{multline}
Here $\eta(u,v,t)$ represents additive noise attributable to ambient background light, detector dark counts, and detector afterpulsing \cite{piron2021}.  We note that the TGSF only accounts for the effects of glare.  It does not account for other effects such as spatial blur due to defocus, or temporal blur due to laser pulsewidth or detector timing jitter.  Hence, if these other effects are present in the observed $y$, they will also be present in the uncorrupted intensity $x$.

In theory, scattering within the receiver optics that produces glare alters the path lengths of measured photons, resulting in a time delay that must be accounted for in the TGSF.  In practice, these time delays are typically very small relative to the time resolution of the lidar sensor, and can be safely ignored.  As a consequence, the TGSF reduces to a \emph{time-independent} glare spread function (GSF), equivalent to the GSF used for glare removal in passive photography \cite{veiling_glare_in_hdr,astronomicalGlare}.  The measurement model can then be updated:
\vspace{-.25cm}
\begin{equation}
    y(u,v,t) = \iint a(u,v,u',v')x(u',v',t)du'dv' + \eta(u,v,t) .
    \label{eq:s_integral}
\end{equation}

\subsection{Correcting Pileup Distortion}
\label{sec:pileup}

\paragraph{Glare and Pileup} The time-independent GSF model of Eq.~\ref{eq:s_integral} suggests a straightforward approach to glare removal: each time slice in the measured datacube can be treated as an independent still image and de-glared separately using photographic de-glare methods \cite{veiling_glare_in_hdr,astronomicalGlare}.  While appealing in theory, this approach fails in practice due to the phenomenon of \emph{photon pileup}, which is inherent to SPAD-based LiDAR and almost always co-occurs with glare in such systems.

\begin{figure}[t]

\vspace*{0.1cm}
\includegraphics[width=0.8\columnwidth]{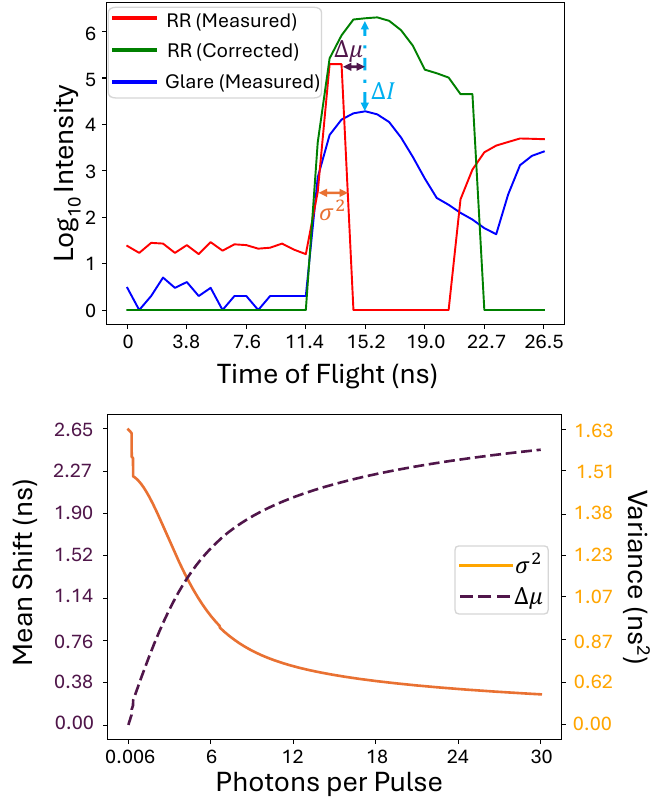}
\centering
\vspace*{-0.3cm}
\caption{(Top) Pileup distortion shifts mean and intensity of histogram corresponding to intense retroreflector (RR) return. RR return induces fainter glare return at a second pixel, which is not affected by pileup. (Bottom) Mean shift and variance of pileup distorted echoes corresponding to a range of incident photon fluxes.}
\vspace*{-0.4cm}
\label{fig:5_pileup_demo}
\end{figure}

Photon pileup occurs because SPADs must switch off for several nanoseconds after each photon detection to reset---a period known as the \emph{dead time} \cite{piron2021}.  When SPADs are subjected to very high fluxes, for which the probability of multiple photon arrivals in a single dead-time period is high, early-arriving photons censor late-arriving photons, distorting the measured pulse shape relative to the true incident waveform.  Glare interference is typically several orders of magnitude fainter than the direct returns that induce it.  As such, problematic glare, in general, only co-occurs with extremely high-flux direct returns from specular, retro-reflective, or very-close range surfaces.

Photon pileup has the practical effects (depicted at top of Fig.~\ref{fig:5_pileup_demo}) of reducing measured flux relative to true incident flux, and shifting the measured signal earlier in time (producing \emph{range-walk error}).  Using Eq.~\ref{eq:s_integral} to naively predict glare from pileup corrupted histograms would yield predictions that were underestimated and temporally misaligned relative to the true glare signal. 

Correcting for pileup at the level of raw histograms can be achieved using Coates's method \cite{coates_correction}, but Coates's method saturates at fluxes as low as 1-10 incident photons per pulse \cite{Heide:2018:subpicosecond}, and as such is inadequate for returns produced by extremely reflective targets like retro-reflectors.  Higher dynamic range pileup correction can be achieved if one estimates the true time of flight and intensity of a finite number of  returns in each histogram \cite{Heide:2018:subpicosecond,rapp2021, sheehan2024} rather than fully ``undistorting'' the raw histogram itself.  For this reason, we argue that glare removal is most effectively addressed among low-level detections---that is, the finite number (e.g.~1, 2 or 3) of returns or \emph{echoes} reported per-pixel, per-frame by many commercial LiDAR sensors.
\smallskip

\noindent \textbf{Moments-based Pileup Correction}
We detect echoes by convolving each raw histogram with a matched filter that is a replica of the transmitted waveform, and then identifying the three highest peaks in the filtered histogram.  We then position a fitting window around each peak and estimate the first three moments of photon arrival times (number of photons, mean time-of-flight, and time-of-flight variance) in each window.  We additionally estimate the level of background noise counts by averaging counts in a noise window that reliably contains only background detections.

If the measured echo intensity is high enough (e.g. $>$~.05 photons per pulse) to produce significant pileup distortion, we estimate the echo's true time of flight and intensity using a moments-based method similar to that proposed by Sheehan et al.\cite{sheehan2024}.  As the true echo intensity increases and pileup becomes more severe, photons bunch up into fewer timing bins---reducing the \emph{variance} of the time-of-flight distribution---and shift to earlier arrival times---shifting the \emph{mean} of the distribution.  We use a photon pileup model to determine the relationship between the true echo intensity, and the measured echo's variance and mean shift.  Our model requires knowledge of waveform shape, detector dead-time, and the rate of background counts.  Moments estimated with the model are shown in \cref{fig:5_pileup_demo}, and theoretical details are provided in the Supplement.

We use the pileup model to compile lookup tables that are queried at measurement time.  First, the true echo intensity is obtained from the measured variance via an inverse lookup.  This estimated intensity is then used to determine the mean shift caused by pileup.  The estimated mean shift is subtracted from the measured mean to obtain the corrected, true mean time-of-flight, which is in turn used to compute the target range.

\vspace*{-0.1cm}
\subsection{Identifying Glare-induced Echoes}
\label{sec:echoid}
\vspace*{-0.1cm}
Our glare suppression method operates in two steps.  First, we use a glare model to predict the expected glare intensity at the time of flight and spatial position of each detected echo.  Second, we compare the predicted glare intensities to the measured echo intensities to determine the likelihood that each echo was induced by glare.  
\smallskip

\begin{figure*}[t]
\centering
\vspace*{0.1cm}
\includegraphics[width=\textwidth]{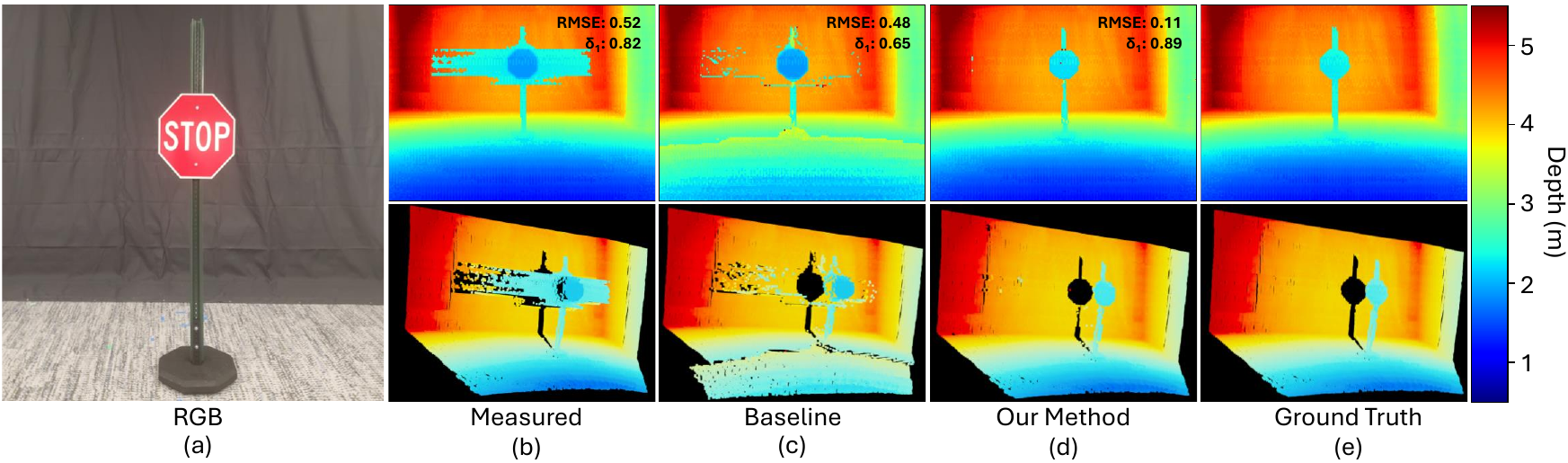}
\vspace*{-0.3in}
\caption{(a) Photograph of captured scene. (b) Measured, glare-corrupted depth. (c) Corrected depth using baseline, deep learning method \cite{deep_learning_bloom_correction}. (d) Corrected depth using our method. (e) Ground truth depth captured by covering retroreflectors with black construction paper. (insets). Root mean square error (RMSE $\downarrow$) in meters and $\delta_1 \uparrow$  (see \cref{eq:delta}) of depth maps relative to ground truth.}
\vspace*{-0.4cm}
\label{fig:3_single_scene}
\end{figure*}

\noindent \textbf{Echo-space Glare Model}
We consider a dToF LiDAR camera with $M$ pixels, and extract the intensity and time-of-flight of $K$ echoes per pixel, per frame.  The vector $\mathbf{y} \in \mathbb{R}_+^{M\times K}$ represents the pileup-corrected intensities (in units of photon counts) of measured echoes, and $\mathbf{t} \in \mathbb{R}_+^{M\times K}$ the pileup-corrected mean times of flight of those echoes.  The vector $\mathbf{x} \in \mathbb{R}_+^{M\times K}$ represents the true intensity of each echo after the influence of glare is removed.

To reformulate the glare model of Eq.~\ref{eq:s_integral} in the discrete echo space, we integrate over the duration of the fitting window used for pileup correction
\vspace{-.15cm}
\begin{align}
\begin{split}
    y_{uk} & = \int\limits_{t_{uk}-\Delta/2}^{t_{uk}+\Delta/2} y(u, t) dt \\ & = \sum\limits_{u' = 1}^M a(u, u')\left[ \int\limits_{t_{uk}-\Delta/2}^{t_{uk}+\Delta/2} x(u',t) dt \right] + \eta_{uk} .
\end{split}
\label{eq:inty}
\end{align}

Here $y_{uk}$ is the measured intensity of echo $k$ at pixel $u$, $y(u,t)$ and $x(u,t)$ are the glare-corrupted and true transient intensities, respectively, defined in Sec.~\ref{sec:transientglare}, and $a(u,u')$ is the GSF.  For compactness, the continuous, two-dimensional spatial coordinates $(u,v)$ have been reduced to a discrete, one-dimensional coordinate $u$.  The symbol $\Delta$ denotes the fitting window duration, and $\eta_{uk}$ is the level of background noise in the fitting window.  Here the window is centered on the echo mean, but it would be straightforward to incorporate a constant offset of the window center.

For a given echo at pixel $u$, the temporal fitting windows corresponding to echoes at \emph{other} pixels may not overlap perfectly in time.  We introduce a \emph{temporal overlap function} $o(\delta t)$---defined as the correlation of the fitting window with the transmitted pulse shape---to quantify the fraction of the signal in another echo $k'$ at pixel $u'$, that contributes glare to echo $k$ at pixel $u$, i.e.
\vspace{-.25cm}
\begin{equation}
    \label{eq:ot}
    o(\delta t) = \int_{-\infty}^{+\infty} \textrm{rect}\left(\frac{t}{\Delta}\right)\lambda(t - \delta t) dt
\end{equation}
where $\delta t = t_{u'k'} - t_{uk}$ and $\lambda(t)$ is the shape of the transmitted waveform, normalized so that $\int_{-\infty}^{+\infty}\lambda(t) dt = 1$.

Eq.~\ref{eq:inty} can now be written in discrete form
\begin{equation}
    y_{uk} = \sum\limits_{u'=1}^M a(u, u')\left[ \sum\limits_{k'=1}^K o(t_{u'k'}-t_{uk})x_{u'k'} \right] + \eta_{uk} .
    \label{eq:gd}
\end{equation}

The above model can be expressed as a linear system $\mathbf{y} = \mathbf{A}\mathbf{x} + \mathbf{\eta}$, where entries of $\mathbf{A}$ correspond to the GSF modified by temporal overlaps.  In theory, it is possible to invert this system to obtain the glare-free intensities $\mathbf{x}$ from the glare-corrupted intensities $\mathbf{y}$.  However the matrix $\mathbf{A}$ depends on measurements of $\mathbf{t}$, which vary frame-to-frame. Inverting a unique linear system for each LiDAR frame would be computationally demanding.  Instead, we approximate the glare intensity $\bar{g}_{uk}$ at pixel $u$, echo $k$ with the operation:
\begin{equation}
    \bar{g}_{uk} = \sum\limits_{u'=1}^M a(u, u')\left[ \sum\limits_{k'=1}^K o(t_{u'k'}-t_{uk})y_{u'k'} \right] ,
    \label{eq:gda}
\end{equation}
which is equivalent to the backprojection $\mathbf{\bar{g}} = \mathbf{\tilde{A}}^T\mathbf{y}$, where $\mathbf{\tilde{A}}$ is equal to $\mathbf{A}$ with diagonal entries set to zero.
\smallskip 

\noindent \textbf{Binomial Likelihood Confidence Metric}
We seek a confidence metric to assess the likelihood that a detected echo contains signal in excess of the predicted glare.  We define $Y$ as the number of photon detections associated with an echo \emph{before} pileup correction.  We assume $Y$ follows a binomial distribution \footnote{This assumption is valid if the fitting window duration is shorter than the SPAD deadtime or, alternatively, if the probability of detecting more than one photon, per laser pulse, in the fitting window is small.}, with the number of trials $N$ corresponding to the number of laser pulses.  We define $P_G = G/N$ as the per trial probability of detecting a photon \emph{if} the incident flux equals the predicted glare flux computed with Eq.~\ref{eq:gda}.  The parameter $G$ is the expected number of \emph{glare} photons expected in $N$ trials\footnote{For moderate to low fluxes, $G\approx g_{uk}$.  For high glare flux where pileup is significant, $G$ must be obtained from an inverse lookup table generated from a pileup model.  See Supplement for details}.  The probability of detecting $Y$ photons in $N$ trials given a per-trial success probability $P_G$ is
\vspace{-.25cm}
\begin{equation}
    \mathbb{P}(Y; P_G, N) = \frac{N!}{Y!(N-Y)!}P_G^Y (1-P_G)^{N-Y}.
    \label{eq:pyg}
\end{equation}
If $Y$ is much greater than $N P_G$, then $\mathbb{P}(Y; P_G, N)$ will be small.  Following this logic, we define the following confidence metric:
\vspace{-.25cm}
\begin{equation}
C=\begin{cases}
			-\ln{\mathbb{P}(Y; P_G, N)}, & \text{if $Y \geq NP_G$}\\
            \quad 0, & \text{if $Y < NP_G$}
\end{cases}
\end{equation}
The confidence metric $C$ acts as a threshold when detected counts lie below the expected counts from glare, and increases monotonically for counts in excess of the predicted glare.

The second-to-last column of Fig.~\ref{fig:teaser} shows a map of confidence values computed for the brightest echo at each pixel, when observing a scene containing multiple retro-reflectors.  We see that, in general, $C=0$ (white pixels) when the brightest echo corresponds to glare, and $C>0$ for non-glare echoes.  In practice, when multiple echoes are detected per pixel, the confidence metric can be used to determine which echoes correspond to real surface returns, and which correspond to glare.
\smallskip\smallskip

\begin{figure}[t]
\centering
\vspace*{0.1cm}
\includegraphics[width=\columnwidth]{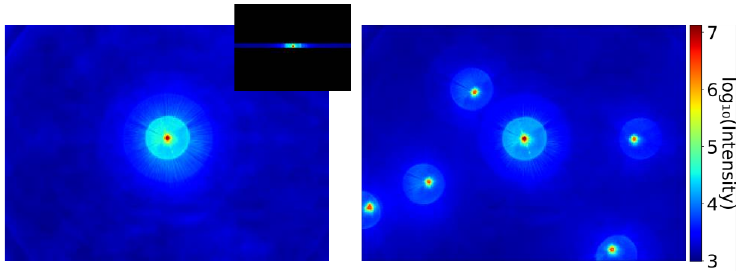}
\vspace*{-0.2in}
\caption{(a) Glare spread function (GSF) captured using infrared flashlight as light source.  Here, source is at FOV center. (inset) Cropped, six-row GSF that accounts for synchronous row-scanning of LiDAR sensor. (b) Superposition of GSFs captured at multiple source positions demonstrates spatial variation of GSF.}
\vspace*{-0.4cm}
\label{fig:2_PSFs}
\end{figure}
\vspace{-.5cm}
\section{Experiments}
\label{sec:experiments}

\begin{figure*}[t]
\centering
\vspace*{0.1cm}
\includegraphics[width=\textwidth]{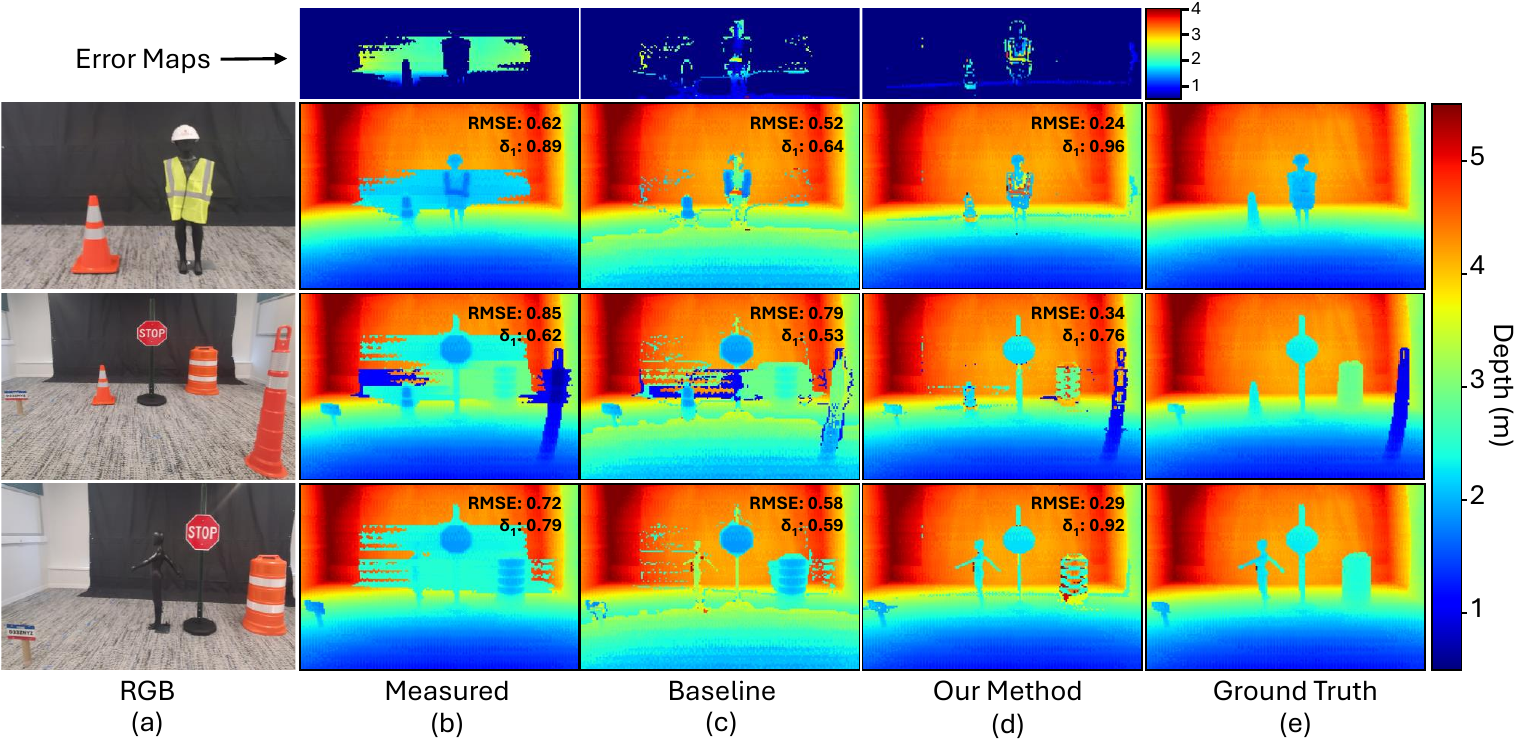}
\vspace*{-0.25in}
\caption{(a) Photographs of captured scenes (b) Measured, glare-corrupted depth maps (c) Depth maps corrected using baseline, deep learning method \cite{deep_learning_bloom_correction} (d) Depth maps corrected using our method (e) Ground truth depth maps captured by covering retroreflectors with black construction paper. (insets) Root mean square error (RMSE $\downarrow$) in meters and $\delta_1 \uparrow$  (see \cref{eq:delta}) of depth maps relative to ground truth}
\vspace*{-0.4cm}
\label{fig:4_multiple_scenes}
\end{figure*}
\vspace*{-0.15cm}

Experiments were conducted with a commercial SP-LiDAR evaluation kit (ADS6311, Adaps Photonics). ADS6311 is a dToF LiDAR array with 192x256 SPAD macro-pixels, that illuminates the scene with a vertically scanned, horizontally oriented line beam.  Evaluations were run on a  RTX~4090 (24\,GB) with Ubuntu~22.04. We demonstrated our method on multiple scenes with varying scene complexity. 

\smallskip 

\noindent \textbf{GSF Acquisition}
To measure the GSF, we turn off the LiDAR transmitter and use an IR flashlight at the same wavelength (940nm). This allowed us to isolate light to a small spot on the receiver array, and alleviated dynamic range constraints imposed by pileup distortion by spreading incident light across all timing bins.  We capture the GSF for 49 light source positions in the LiDAR FOV, and use a normalized distance-weighted sum to interpolate to light source positions that were not measured.  An example GSF with the light source at the FOV center is shown in Fig.~\ref{fig:2_PSFs}.

ADS6311 scans the scene vertically, switching on groups of six pixel rows in sync with the scan position of the transmitted beam.  This means that during LiDAR measurements, glare is observed in six-row bands around the aggressor pixel.  The banding is not observed in flashlight measurements because the flashlight is not scanned synchronously with sensor row activations.  We re-create the banding effect by masking pixels outside a band that contains the light source (inset, Fig.~\ref{fig:2_PSFs}).

\smallskip

\noindent\textbf{Evaluation Setup}
To evaluate our method we used a number of different retroreflectors common on the road: a stop sign, license plate, traffic cones, high-visibility traffic safety vest and a traffic drum. The per-frame exposure time was set low to ensure points on the retroreflector remained under the count limit (4096 counts per timing bin per frame). Because of the ubiquity of roadsigns, our first experiments conducted were on a stop sign. The results of this test can be found in \cref{fig:3_single_scene}
\smallskip

\noindent\textbf{Ground Truth}
For quantitative results, we collected the ground truth by masking the retroreflective elements with a combination of black construction paper and black tape. To our knowledge, no extensive glare datasets for LiDARs exist. One of the challenges to collecting such a dataset is acquiring ground truth data.  One of the benefits of a non-learning based approach to glare correction is we do not need a large training set to achieve robust performance. Self-supervised approaches do not require ground truth data but they do require large training sets.
\vspace{1pt}

\noindent\textbf{Comparison to Deep Learning Method}
Because there are limited academic works addressing glare suppression, we select a representative method \cite{deep_learning_bloom_correction} for comparison. \cite{deep_learning_bloom_correction} uses depth and intensity maps to model the GSF as an exponentially decaying function and uses synthetic data for training. The method suppresses glare artifacts in the depth map space. Because the training paradigm requires simple retroreflector geometry (e.g. octagons or rectangles) we observed this approach struggled on complex geometries such as the cone or high-vis vest. Scene independence is an important benefit of our method, which learning-based approaches lack without an extensive training dataset.

\smallskip

\begin{figure*}[t]
\centering
\vspace*{0.1cm}
\includegraphics[width=\textwidth]{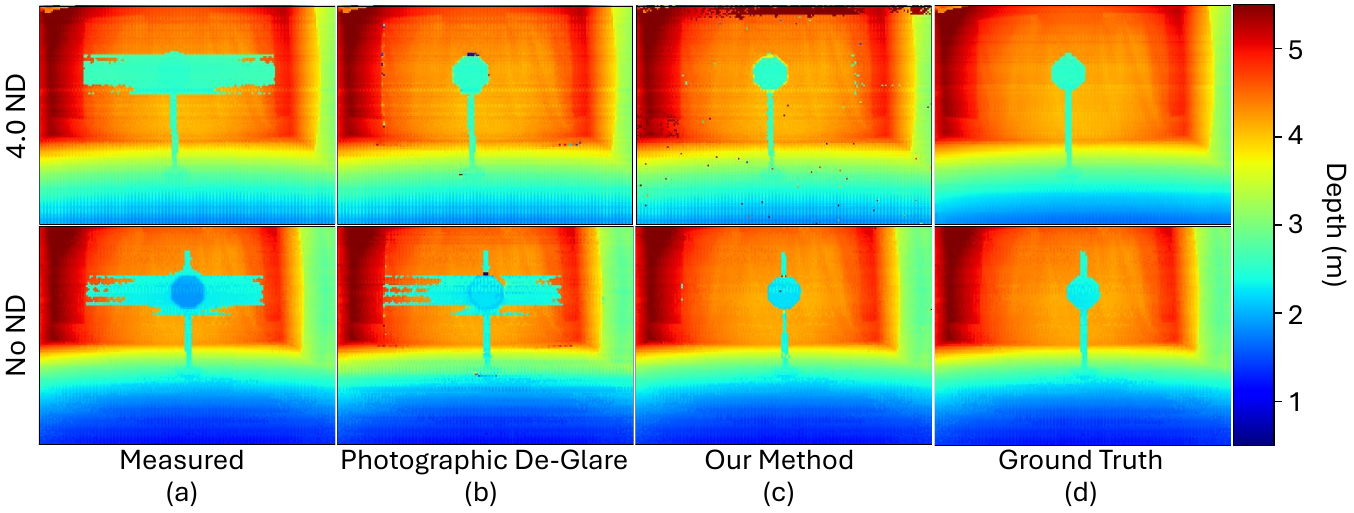}
\vspace*{-0.3in}
\caption{Comparison of our method to a photographic de-glare method that uses an algorithm proposed by Talvala et al.~\cite{veiling_glare_in_hdr} to independently remove glare from each datacube time slice, as if it were a still image. (top) When pileup is mitigated by attenuating the signal with a neutral density filter (NDF, OD4), both methods mitigate glare successfully. (bottom) Without the NDF, severe pileup causes the photographic de-glare approach to fail.}
\vspace*{-0.4cm}
\label{fig:6_NDFilterTest}
\end{figure*}
\vspace{1pt}

\noindent \textbf{Aggregate Scene Results}
To demonstrate the generality and versatility of our approach, we show results on scenes with multiple retroreflectors of various types. For one scene we placed retroreflectors of varying size and shape at different depths. In another scene we emulated a scenario of a child crossing the road. For this we used a black mannequin which offered a challenging test case as weaker signals are more difficult to recover. We placed the mannequin at the same depth as the stop sign and a traffic drum to demonstrate that the signal can still be recovered when masked by bloom corruption in the same timing bins. We also show that retroreflector elements when worn on the person (in the form of a high-vis traffic safety vest) can still be suppressed while retaining the signal. This is an important component of our method because other glare suppression techniques are too aggressive and can remove the signal in addition to the corrupted region. This is problematic if a person is hidden by the glare. The qualitative results of these evaluations are shown in \cref{fig:4_multiple_scenes} with quantitative results inset.  For quantitative analysis we report  $\delta_1$ and RMSE. $\delta_i$ is given by: 
\vspace{-.25cm}
\begin{equation}
\label{eq:delta}
\delta_i = \frac{1}{N} \sum_{j=1}^N \mathbf{1}(\max(\frac{d_j}{\hat{d_j}},\frac{\hat{d_j}}{d_j}) < (1+\frac{i}{100}))  
\end{equation}
where $\mathbf{1}$ is the indicator function, $N$ is the number of pixels, and $d_j,\hat{d_j}$ is the ground truth depth and predicted depth at the $j$-th pixel respectively. Because our method works in the echo space it can be easily integrated into existing LiDAR DSP pipelines. 

\noindent \textbf{Comparison to Photographic De-glaring Method}  Here we demonstrate how photon pileup causes photographic de-glaring methods to fail when applied to individual time slices of a raw SP-lidar datacube.  We capture two sets of measurements for the scene containing a retroreflective stop sign.  In the first, data is captured normally and histograms associated with pixels containing the stop sign show severe pileup distortion.  In the second, a neutral density filter is placed in front of the lidar transmitter to reduce transmitted intensity by a factor of $10^4$, which reduces the intensity of retroreflector returns such that pileup is negligible. The integration time of the second measurement is simultaneously increased (to 1.16 seconds), so that the total number of detected photons in both measurements are comparable.

For both sets of measurements, we treat each time slice of the captured datacube as an independent still image and attempt to remove glare from each time slice using a single-step glare removal operator\footnote{See Supplement for theoretical details.} proposed by Talvala et al.~\cite{veiling_glare_in_hdr}.  We then employ a typical peak-finding procedure on the ``de-glared'' datacube, and form a depth map by selecting the depth of the brightest peak at each pixel.  In \cref{fig:6_NDFilterTest} we compare depth maps obtained this way to depth maps produced with our method, which instead selects the depth of the highest confidence echo.  Both methods successfully suppress glare when there is no pileup.  However, in the case of the unmodified sensor, where pileup is severe, only our method, which operates on echos rather than the raw datacube, successfully suppresses the glare.  The failure of the photographic de-glare method is due primarily to range-walk error induced by pileup.  Pileup shifts direct returns from glare to earlier time bins, while the fainter glare signal is not shifted in time.  This separates glare and aggressor signals into different time slices, such that linear de-glare methods operating on individual time slices over-predict glare in earlier time slices, and fail to remove the true glare.

\vspace{-.7cm}
\section{Limitations and Future Outlook}
\label{sec:disc}
\vspace{-.15cm}
One limitation of our work is that we have yet to test our method in outdoor traffic scenes where, in particular, solar background is more intense. Our method also requires pre-calibration of a GSF unique to each sensor and the validity of this GSF may degrade in realistic conditions---e.g. dust, condensation, or scratches on optical elements.

Looking forward, we expect de-glare to become a standard component of 3D perception pipelines—analogous to denoising or compression in conventional imaging. Historically, legacy LiDARs adopted suboptimal spinning architectures to avoid glare. Real-time algorithmic de-glare could reduce the need for such constraints, freeing designers to pursue fully solid-state, high-speed, and high-resolution architectures. The ability to correct glare algorithmically thus expands the design space for the next generation of “camera-fied” LiDARs.

\section{Acknowledgments}
\label{sec:acknowledge}
This research was supported by NSF CAREER Award 1943149, NSF grant CNS-2107060, NRT Award Number 2152163, and ONR grant N000142412155. We thank Sacha Jungerman for helpful discussions.


{
     \small
     \bibliographystyle{ieeenat_fullname}
     \bibliography{main}
}
\clearpage
\setcounter{page}{1}
\maketitlesupplementary

\label{suppl:additional_evaluation_details}

\section{Implementation Details}

We captured scenes at a per-frame exposure of 1.16$\mu$s for 50 frames. The reason for capturing many frames with a low exposure is a constraint on the dynamic range of our sensor due to photon count overflow. The ADAPs sensor used in this study cannot store more than 12 bits of photon counts per time bin, per exposure.  If more photons are detected in a single time bin, the measured histogram will be clipped to this maximum value. This poses an issue because during capture of scenes with severe pile-up we will have ambiguity with points that exceed this limit and we cannot accurately measure variance of time of flight as required by our pile-up correction method.

\label{suppl:GSF}

\section{Glare Spread Function Details}

\begin{figure*}
    \centering
    \includegraphics[width=\linewidth]{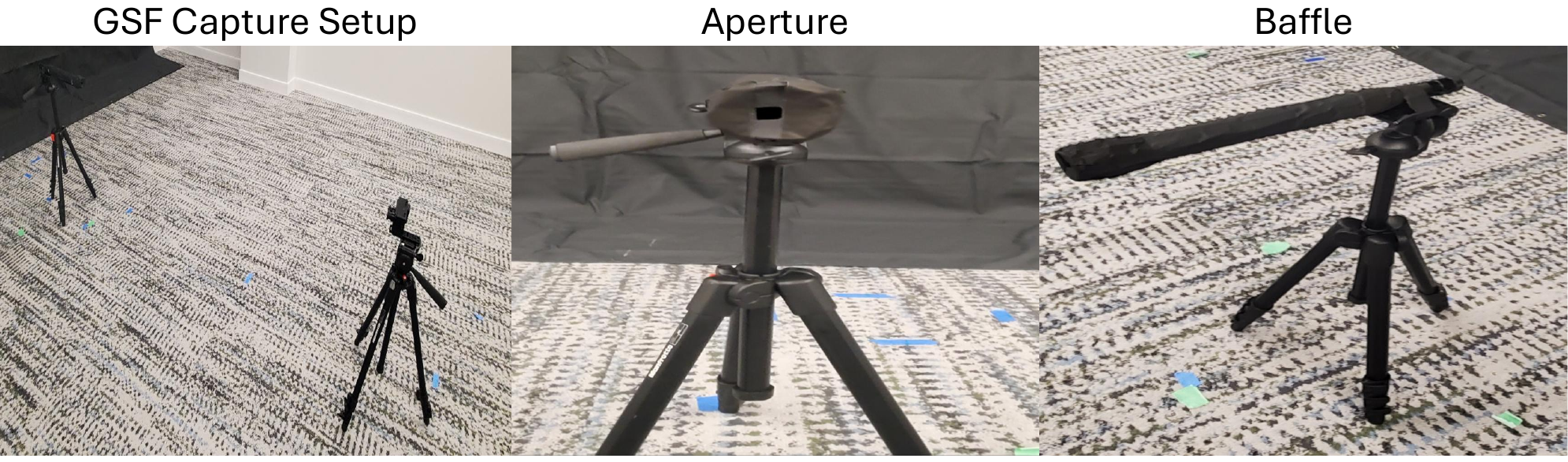}
    \caption{Our setup for capturing the GSFs. We include an image of the full setup as well as closeups of the baffle and aperture size. This was recreated but is roughly the same distance apart as during the GSF captures. The same baffle and aperture was used during capture.}
    \label{fig:GSF_capture_setup}
\end{figure*}

\begin{figure*}[t]
\centering
\vspace*{0.1cm}
\includegraphics[width=\textwidth]{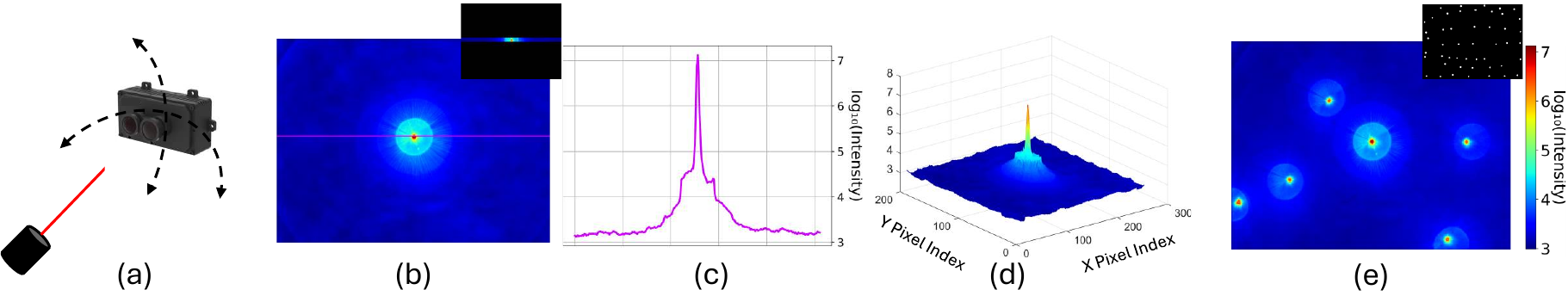}
\vspace*{-0.3cm}
\caption{(a) Animation depicting our GSF acquisition setup. We used an IR flashlight with a small aperture and baffle to generate bright points in the LiDAR field of view (FOV). To measure different locations we kept the flashlight at a fixed location and pivoted the LiDAR. 
(b) A single measured GSF positioned at the center of the LiDAR FOV. The cropped inset depicts a horizontal band of the GSF. In practice this band was used as a GSF to model the behavior of the scanning system of the LiDAR. 
(c) Horizontal cross section of the GSF depicted in (b). Intensity is taken after background subtraction. 
(d) Log-scale surface plot of GSF. 
(e) Depiction of the spatial variation in GSF shape.  We recorded GSFs at 49 locations in the LiDAR FOV (all locations shown in inset).}
\vspace*{-0.4cm}
\label{fig:GSF_expanded}
\end{figure*}

\paragraph{Measurement} To measure the GSF, we turned off the laser of our LiDAR and used an IR flashlight at the same wavelength (940nm). This allowed us to avoid the quantization limit of our sensor by distributing the energy over time, giving us a more accurate measure of the GSF. It additionally allowed us to remain well under the saturation limit of our sensor and avoid non-linear behavior. By using a baffle on the end of our flashlight, we could isolate the internal multipath bounces and avoid noise from external multipath bounces in the measurement environment. A small aperture on the end of the baffle was used when measuring the GSF to ensure that light from the flashlight was imaged onto at most one pixel in the array. This also ensured that the intensity was low enough to not completely oversaturate our sensor (causing pixels to ``shutoff"). To capture any spatial variation of the GSF, we repeat this process to capture GSFs at 49 locations within the LiDAR FOV. The setup for capturing the GSFs can be found in \cref{fig:GSF_capture_setup}. 

\paragraph{Processing} An analysis of our GSF characterization efforts is shown in \cref{fig:GSF_expanded}. Because our LiDAR is 1D vertical scanning (with six rows of pixels switched on at a time), there is a correlation between the transmitted pulse and the received transient; this is why blooming artifacts appear as horizontal bands spanning six rows. Our method for measuring the GSF does not produce this pattern because the light is constant (not pulsed), and not scanned in sync with row activations in the receiver array. Thus, to mimic the effect of the scanning mechanism we only used a six-pixel wide horizontal band of the measured GSFs centered around the aggressor as depicted in \cref{fig:GSF_expanded}. To normalize the GSF consider the unnormalized (measured) GSF $\mathbf{a}$ and let $N = \sum_i a_i$ be the total number of photon counts in $\mathbf{a}$ and $\alpha$, the outscatter ratio, be defined as: 
\begin{equation}
    \alpha = 1 - \frac{a_0}{N}
\end{equation}
where $a_0$ is the number of photon counts in the pixel that contains the GSF peak. The normalized $GSF$ is given by: 
\begin{equation}
    \hat{\mathbf{a}} = \begin{cases}
    0 & \text{at GSF peak pixel} \\
    \frac{a_i}{\alpha N}& \text{elsewhere}
\end{cases}
\end{equation}
We further weight the GSF by a distance dependent factor 
\begin{equation}
    e^{w||p_0 - p_1||_2}
    \label{eq:exp_decay_distance}
\end{equation} where $p_0$ is the coordinate location of the GSF peak pixel and $p_1$ is the pixel being weighted. We set $w=.01$. 

\label{suppl:DSP}
\section{DSP Pipeline}
\begin{figure*}[t]
\centering
\vspace*{0.1cm}
\includegraphics[width=\textwidth]{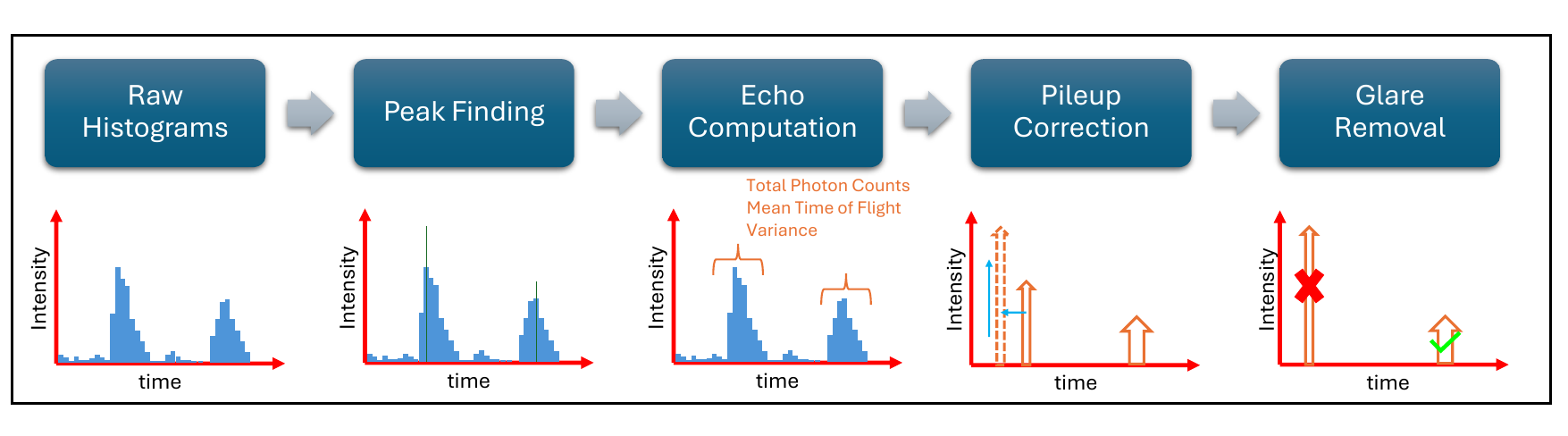}
\vspace*{-0.3cm}
\caption{Our DSP pipeline is typical of one likely found in commercial SP-LiDAR devices. Starting with raw histograms, peak finding is done to determine echos. We then use the total photon counts, mean time of flight and time of flight variance of a fitting window around the peak to perform pileup correction in the echo space. Finally, we perform glare removal as described throughout the paper.}
\vspace*{-0.4cm}
\label{fig:signal_processing}
\end{figure*}

Our approach follows a typical digital signal processing pipeline that might be found in a commercial SP-LiDAR (though these are typically proprietary so details may vary).  An overview of our pipeline can be found in \cref{fig:signal_processing}. The details of peak finding and echo computation is given below.

\paragraph{Peak Finding}
We first locate peaks by taking the top $n$ highest count bins after convolving the signal with a kernel. This kernel was measured from a scene containing a flat, low-reflectance surface to avoid nonlinearities in the measurement. 
\paragraph{Echo Computation}
We take these $n$ peaks and compute the first three moments around some fitting window: total measured counts, mean time of flight, and variance of the time of flight. With these we can start to consider pile-up. 
\label{suppl:LUT_pileup_correction_theory}

\begin{figure*}
\vspace*{0.1cm}
\includegraphics[width=\textwidth]{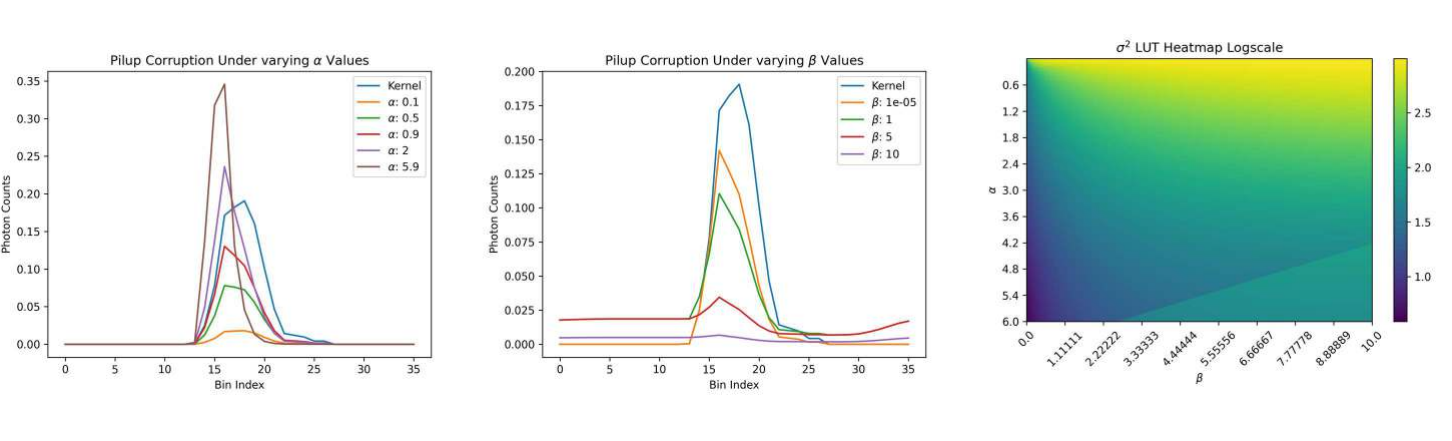}
\vspace*{-0.3cm}
\caption{This demonstrates the forward model as alpha and beta are varied and gives a heatmap for the variance lookup table.}
\vspace*{-0.4cm}
\label{fig:forward_model}
\end{figure*}

\begin{figure*}
    \centering
    \includegraphics[width=\textwidth]{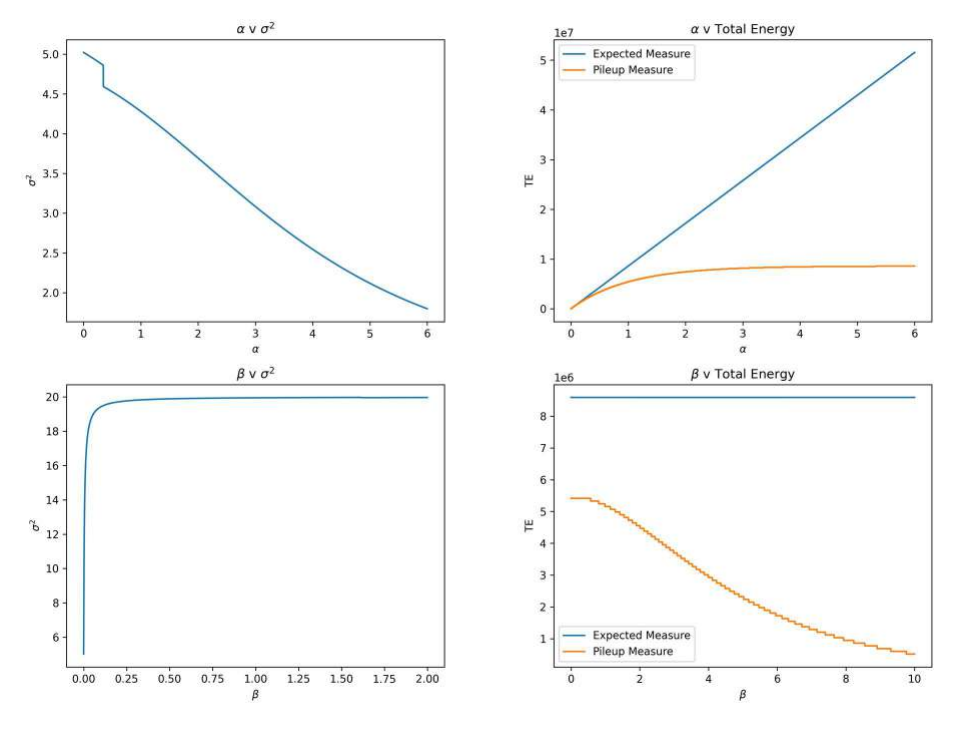}
    \caption{The top row shows the change of total energy and variance as alpha is varied. Two things to note here are that as alpha goes up, variance decreases. The same happens with total energy, we expect linear behavior as in the blue curve, but due to pileup and dead-time effects we get nonlinearities as depicted in the orange curve. Our pileup correction method recovers this linear behavior. For the bottom row variance increases as background counts begin to dominate the signal and energy decreases due to dead-time effects. Photon counts in the we are computing the energy on are lost due to background photons causing a deadtime to occur before the signal reaches the sensor.}
    \label{fig:alpha_beta_trends}
\end{figure*}

\section{Pileup Correction}
Photon pile-up occurs when incident flux is too high, leading to photon arrivals within a SPADs deadtime that are not detected. This is a non-linear effect which poses a significant issue because our method assumes a linear relationship between the intensity of incident light and the number of measured photon counts.
\subsection{The Problem of Pileup Distortion}

Retroreflectors cause extreme pile-up because of the intensity of the return. The glare component, being significantly weaker, does not see the same pile-up. 

Pile-up correction is a topic of current research and there are methods to correct for it. Coates correction \cite{coates_correction} is one common method to address pile-up. Another more recent method can be found in \cite{pileup_correction}. The issue with both of these methods is they only correct pile-up in the scenarios where some photons are detected in later bins, they assume there is a bias that shifts the peak, but some parts of the signal in bins later than the peak still exist. In the scenario where pile-up is so extreme that no photons appear in later bins, these methods do not work. 

One way to avoid extreme pile-up and further mitigate the overflow saturation as previously mentioned is to use a neutral density filter. However, we found to effectively mitigate pile-up effects with this approach we required an extremely strong filter with a strength of OD 4.0 to attenuate the signal. This is not practical in a real deployment scenario as the signal is too weak in even moderately noisy conditions to rise above the noise floor. Thus, we provide a method to mitigate pile-up that utilizes a pile-up corruption forward model to populate lookup tables which can then be used to correct the range-walk errors and intensity underestimation in extreme pile-up conditions.

\subsection{Correction Method}
To build these lookup tables lets first discuss the forward model for pileup corruption. 
\paragraph{Forward Model}
Consider some normalized transmitted signal $\vec\lambda^s \in \mathbb{R}^T$ for $T$ timing bins such that $\sum_i^T \vec\lambda_i^s = 1$ with some background noise $\vec{\lambda}^b \in \mathbb{R}^T$ where $\lambda_i^b = \frac{1}{T}$, i.e. $\sum_i^T \lambda_i^b = 1$. Thus the incident signal is just a linear combination of the transmitted signal with the background:
\begin{equation}
    \vec\lambda = \alpha \vec \lambda^s  + \beta \vec \lambda^b = \alpha \vec \lambda^s + \frac{\beta}{T}
\end{equation}
where $\alpha$ is the number of signal photons detected and $\beta$ the number of background photons detected \textit{\textbf{per-pulse}} when summed over all time bins. 

Because counts follow a poisson distribution under the assumption there is no pileup we can say that the probability of detecting $\geq 1$ photon in bin $i$  is $p_i = 1 -e^{-\lambda_i}$ where $\lambda_i = \alpha \vec \lambda_i^s + \frac{\beta}{T}$. Thus, if we consider the effect of some dead-time $D$ (given in number of bins) we can compute the $\mathbb{E}[\text{detections in bin }i]$ per pulse by:
\begin{equation}
q_i(\alpha,\beta,\lambda) = p_i
\prod_{\mathclap{\scriptstyle j=(i-D-1)\bmod T}}
      ^{\mathclap{\scriptstyle (i-1)\bmod T}}
(1 - p_j) = (1-e^{-\lambda_i}) \prod_{\mathclap{\scriptstyle j=(i-D-1)\bmod T}}  ^{\mathclap{\scriptstyle (i-1)\bmod T}} e^{-\lambda_j}
\label{eq:pileup_forward}
\end{equation}
 To improve the numerical stability we compute the logarithm of \cref{eq:pileup_forward} as:
\begin{equation}
    \ln[q_i(\alpha,\beta,\lambda)] = \ln(1-e^{-\lambda_i}) - \sum_{\mathclap{\scriptstyle j=(i-D-1)\bmod T}}
      ^{\mathclap{\scriptstyle (i-1)\bmod T}} \lambda_j
    \label{eq:stable_pileup_forward}
\end{equation}

\paragraph{Generate Lookup Tables}
We then create lookup tables for the 0th, 1st and 2nd moments of a pileup corrupted signal which we will refer to as $LUT_{\gamma}$ $LUT_{\mu}, LUT_{\sigma^2}$ respectively. To do this first we start with a pileup free signal, $\lambda_k$, which we measured using a dark background. 

We then compute $\hat q_i = q_i(\alpha,\beta,\lambda_{k,i})$, the pileup corrupted waveform, for evenly spaced $\alpha \in [0,a]$ and $\beta \in [0,b]$ and compute the LUT entries as $LUT_\gamma[\alpha,\beta] = \sum_j \hat{q}_j$ $LUT_{\mu}[\alpha,\beta] = \mu_\lambda - \mu_{q}$ where $\mu_q$ and $\mu_\lambda$ are the means computed over some fitting window $j$ around the peak of the pileup corrupted waveform and the representative waveform (kernel) respectively. $LUT_{\sigma^2}[\alpha,\beta] = \sigma_{\hat q}^2$ where the variance is computed over some fitting window around the peak of the pileup corrupted waveform. See \cref{fig:forward_model} for the effects of varying $\alpha$ and $\beta$ as well as a heatmap of the $LUT_{\sigma^2}$. See \cref{fig:alpha_beta_trends} for a more detailed view of the $LUT_{\sigma^2}$ and $LUT_\gamma$ under varying $\alpha$ and $\beta$.
\begin{figure*}
    \centering
    \includegraphics[width=0.6\textwidth]{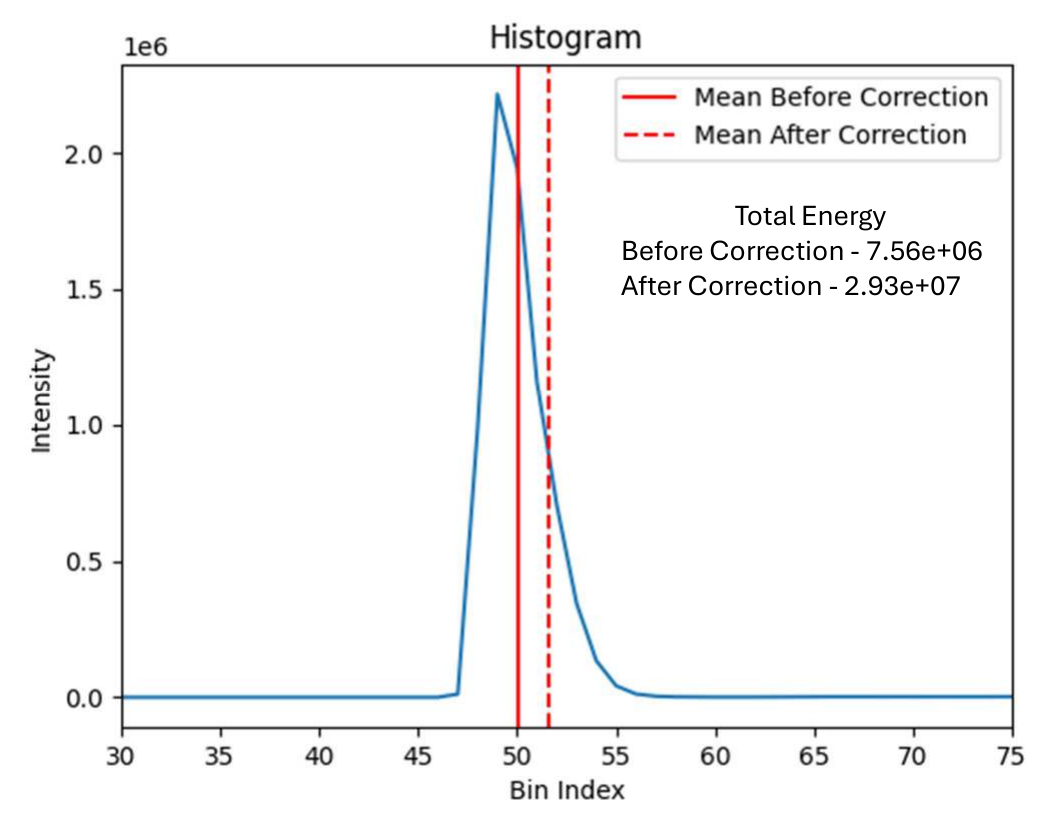}
    \caption{This shows the histogram under pileup corruption from a real capture with a stop sign. This is still with a relatively strong ND filter, with no ND filter the pileup may be too high for this method to work. You can see though that there is a mean shift and energy correction to this pileup corrupted waveform.}
    \label{fig:histograms}
\end{figure*}
\paragraph{Pileup Correction}
Once we have these lookup tables we can correct for pileup in a measured signal $\lambda_m$. First, consider a typical signal processing pipeline for SP-LiDARs where we would first perform some kind of peak detection and get $n$ peaks/pixel (referred to as ``echos'' in the literature). For each of these echos we can compute the statistics (total energy, variance and mean) of some fitting window around the peak (typically the same size as what is used in the LUT creation) to get some $\gamma_{m}, \sigma^2_{m},\mu_{m}$. 

\begin{figure*}
    \centering
    \includegraphics[width=\linewidth]{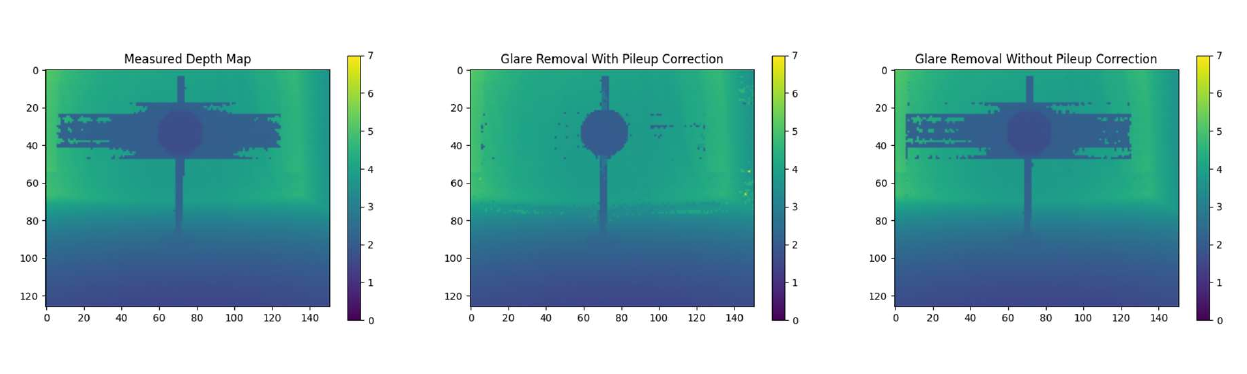}
    \caption{Here is a demonstrated use case of this method. The middle image depicts glare correction without pileup correction and the right image with pileup correction. As you can see points around the stop sign are not corrected due to the distortion caused by pileup.}
    \label{fig:pileup_comparison}
\end{figure*}

We compute the average measured background $\beta_m$ as the mean of the last 557 bins (note that with the limited depths in the scene these bins only contained background counts). With $\beta_m$ and $\sigma^2_m$ we can first find the corresponding row in our lookup table $LUT_{\sigma^2}[:,\beta_m]$ and find the closest fitting $\alpha$ (which we will call $\alpha_m)$ using interpolation or some search method like binary search. With $\beta_m,\alpha_m$ we can then correct the mean shift (\cref{fig:histograms}) by 
\begin{equation}
\mu_m' = \mu_m + LUT_\mu[\alpha_m,\beta_m] = \mu_m + \mu_\lambda - \mu_q  
\end{equation}
 and the total energy by 
 \begin{equation}
     TE_m' = \alpha \cdot (\# \text{pulses}).
 \end{equation}
 Doing this for each of the $n$ echos gives us $n$ pileup corrected echos and fits into traditional DSP pipelines. We will also use $\gamma' = LUT_\gamma[\alpha_m,\beta_m]$ for computing expected glare. See \cref{fig:pileup_comparison} for a comparison of our method with and without pileup correction.

\section{Our Method}
Once we have echos that are pile-up corrected we can move to the glare removal part of our pipeline.

\paragraph{Band Size}
While in theory (according to provided documentation) a 6 row band would be optimal, we found that in practice a larger band size worked better so we extended this to 17 rows.

\paragraph{EigenCWD Details}
To address the spatial variance of the GSFs, one could naively interpolate the remaining GSFs, this however would require a massive 49152x49152 array which would be computationally expensive. To get around this we took inspiration from \cite{eigenCWD}. Instead of interpolating the array, we store the weights (computed by the inverse square of this distance to an unmeasured point) for each of the 49 GSFs giving us a much more compact 49152x49 size array. We then computed the expected glare at some pixel $u,v$ coming from another pixel at $u',v'$ using a normalized weighted sum of the measured GSF, $\widehat{GSF}$, as \begin{equation}
    \mathbb{E}[G_{(u,v)}] = \widehat{GSF}_{(u',v')} \cdot \gamma'_{(u',v')}
\end{equation} where $\gamma'$ is from \cref{suppl:LUT_pileup_correction_theory}

Because this weighted average requires shifting the GSFs, the borders may be unmeasured, to account for this, rather than zero padding, we expanded the borders using the same distance weighted exponential as in \cref{eq:exp_decay_distance} with the base value determined by the pixels at the edge of the pad region. 

\paragraph{Background Determination}
We computed the background the same as in \cref{suppl:LUT_pileup_correction_theory} but we further set a minimum background of 53 photons to account for higher background counts in earlier bins due to effects such as afterpulsing.

\paragraph{Depth Determination}
After computing the negative log liklihood, one may notice that echos with measured counts well under the expected glare are treated similarly to points with measured counts well above the expected glare. To solve for this we zero out points that are below the expected glare. Because we also zero out points that are below 5$\sigma$ of the expected background, in the event that all peaks are zero'd out, we instead retain peaks that are below the expected glare and weight them by the sigmoid of $\frac{\text{counts}_{measured} - \mathbb{E}[Glare]}{T}$ with $T=90$. The echo with the highest confidence is then selected as the depth.



\section{Retroreflector Analysis}

Early on in this work we examined how different retroreflector characteristics effected glare. We looked at retroreflector sizes, distances and colors. 
\paragraph{Size}
The size of the retroreflector directly contributes to the extent of the glare. Each pixel that contains a retroreflector element not only contributes to the glare of pixels outside the retroreflector but also other adjacent pixels on the retroreflector. This is part of the reason pile-up was so problematic. In theory, every scene element would contribute some glare, in practice, elements that do not cause strong returns have a low enough probability of causing glare that it is usually not detected. However, even retroreflective elements that spanned a single pixel we noticed caused some glare in adjacent pixels demonstrating the high intensity of these returns. 
\paragraph{Distances}
Distance is less interesting, as it simply follows the inverse square law of intensity falloff. But again, distance is important when considering the need to mitigate pile-up. 
\paragraph{Color}
We tried multiple colors of retroreflectors (yellow, red, orange, white) and multiple colors were demonstrated in our results. We found color had negligible effects and thus we did not perform an ablation study for this. 
\label{sec:suppl_Ablations}

\section{Photographic De-glare Method}

\subsection{Pileup-free Measurement Model} When pileup is minimal, measurements collected by a LiDAR with a camera-like receiver are well modeled by Eq.~2 of the main text. If the GSF is known, the uncorrupted datacube $x$ can be recovered from measurements $y$ by solving a linear inverse problem.  Furthermore, because the GSF is effectively time-independent, we can solve for each time slice of the datacube independently.  This allows us to reduce the large problem of inferring a three-dimensional data volume from three-dimensional measurements, to a set of smaller, two-dimensional image recovery problems.

We write down a discrete version of Eq.~2 in the main text:

\begin{equation}
    \mathbf{y}_t = \mathbf{A}\mathbf{x}_t + \mathbf{\eta}
    \label{eq:discrete_model}
\end{equation}

Here $\mathbf{y}_t$, $\mathbf{x}_t \in \mathbb{R}^{MN}$ are $M\times N$ pixel images that correspond to time slices of the measured and uncorrupted datacubes, respectively, at time $t$. $\mathbf{\eta}\in\mathbb{R}^{MN}$ is a random noise vector, which we assume follows a noise distribution that is independent of $t$. The GSF is represented by the matrix $A\in\mathbb{R}^{MN\times MN}$.

\subsection{Fast Glare Removal with Sharpening Operator}
\label{sec:sharpen}

We use a single-step glare removal method originally proposed by Talvala et al.~\cite{veiling_glare_in_hdr} for glare removal in high dynamic range photography.  We use the method to remove glare from individual time slices of a LiDAR datacube, which we treat as still images.  The single-step glare removal method is similar in spirit to sharpening operators used in image processing \cite{szeliski_chapter3}---that is, an approximate inverse that can be applied to an image in minimal computation time.   Although iterative optimization methods in theory provide accurate solutions, they are arguably too computationally intensive to provide real-time de-glaring in such scenarios.  While the quality of de-glared images produced using single-step de-glare operators may suffer relative to those produced using iterative optimization, our ultimate goal is not to produce clear images, but rather to suppress glare signals that might otherwise be interpreted as real objects in the scene.  We will show that, when pileup distortion is not severe, our approximate de-glare operators are effective at accomplishing this task.  

To construct the de-glare operator, we first deconstruct the GSF into two components: an identity matrix that accounts for unscattered light, and a second operator $\mathbf{B}$ that accounts for light scattered away from the intended pixel:

\begin{equation}
    \mathbf{A} = (1-\alpha)\mathbb{I} + \alpha \mathbf{B} .
    \label{eq:deconstructA}
\end{equation}

Here $\alpha$ quantifies the fraction of light scattered away from the uncorrupted image.\footnote{For notational simplicity, we've assumed the fraction of outscattered light is the same for all image pixels.  To account for variable outscatter fractions, we could define a vector $\mathbf{\vec{\alpha}}$ of same length as $\mathbf{x}_t$, and replace Eq. \ref{eq:deconstructA} with $\mathbf{A} = \mathrm{diag}[\mathbf{1} -\mathbf{\vec{\alpha}}] + \mathbf{B}\ \mathrm{diag}[\mathbf{\vec{\alpha}}]$ }, with $0 \leq \alpha \leq 1$.  All entries of $\mathbf{B}$ are positive, and all columns of $B$ sum to one--i.e. $\sum_i B_{ij} = 1$ for all columns $j$--which ensures that $A$ conserves the energy of the uncorrupted image.

We construct the following glare removal operator $\mathbf{S}$ from components of the GSF, and apply it to measurements $\mathbf{y}_t$ to obtain the approximate solution $\mathbf{\hat{x}}_t^{(S)}$:

\begin{equation}
    \label{eq:deglarematrix}
    \mathbf{\hat{x}}_t^{(S)} = \mathbf{S}\mathbf{y}_t = \left[ (1+\alpha)\mathbb{I} - \alpha \mathbf{B}\right]\mathbf{y}_t .
\end{equation}

The operator $\mathbf{S}$ subtracts the unwanted glare component while simultaneously scaling the input image to account for light lost due to outscatter.  Like $\mathbf{A}$, the columns of $\mathbf{S}$ sum to 1, and $\mathbf{S}$ is energy-conserving.  The errors in  $\mathbf{\hat{x}}_t^{(S)}$ scale with $\alpha^2$, and so we expect $\mathbf{\hat{x}}_t^{(S)}$ to be a reasonable approximation when $\alpha$ is small.  Importantly, $\mathbf{\hat{x}}_t^{(S)}$ can be evaluated with a single matrix-vector product, which requires significantly less computation time that an iterative solver.

The evaluation of $\mathbf{\hat{x}}_t^{(S)}$ can be sped up further by assuming that the GSF is \emph{shift-invariant}, which is often approximately true--particularly at the center of the lidar field of view.  In this case, the matrix $\mathbf{B}$ becomes the convolutional kernel $\mathbf{b}$, and the de-glare operator can be written as

\begin{equation}
\label{eq:deglare_operator}
    \mathbf{\hat{x}}_t^{(S)} = (1+\alpha)\mathbf{y}_t - \alpha \left(\mathbf{b}*\mathbf{y}_t\right)
\end{equation}

\subsection{Experimental Details}

For evaluating the PDG method we needed to use an ND filter, however, because the lens on our ADAPs evaluation kit is not replaceable we needed to mount the filter over the lens. It is worth noting that we mounted the ND filter in front of the transmitter, not the receiver, as mounting it in front of the receiver would alter the GSF that was measured without use of the ND filter. The filter caused significant crosstalk between the transmitter and receiver, so we used a crosstalk barrier. The ADAPs kit comes with a crosstalk barrier which we extended with a small piece of black foil.

This foil was within the FOV of the sensor thus when applying any method we first zeroed out the borders so this did not impact the glare removal process. Additionally, we cropped out these regions for the figures. We used the same cropped region for the scenes with multiple retroreflectors.

\subsection{De-glare Operator Computation}
To generate the convolutional version of the de-glare operator described in Eq.~\ref{eq:deglare_operator}, we use only one GSF, captured at the center of the lidar FOV (see Fig.~\ref{fig:GSF_expanded}b,c,d).  The unnormalized GSF $\mathbf{a}$ can be decomposed as

\begin{equation}
        \mathbf{a} = N\left[  (1-\alpha)\hat{e}_0 + \alpha \mathbf{b}\right] ,
\end{equation}

where $N = \sum_i a_i$ is the total number of photon counts in the measured GSF, $\hat{e}_0$ is a one-hot vector that equals 1 at the pixel that contains the GSF peak and 0 elsewhere, $\mathbf{b}$ is the convolutional kernel described in Eq.~\ref{eq:deglare_operator}, and $\alpha$ is the outscatter ratio from Eq.~\ref{eq:deglare_operator}.

The outscatter ratio $\alpha$ can be computed as follows:

\begin{equation}
    \label{eq:compute_alpha}
    \alpha = 1 - \frac{a_0}{N} ,
\end{equation}

where $a_0$ is the number of photon counts in the pixel that contains the GSF peak.  The convolutional kernel $\mathbf{b}$ can be computed by taking the vector $\mathbf{a}/N$ and setting the entry of the GSF peak pixel to zero, i.e.

\begin{equation}
b_i =
\begin{cases}
    0 & \text{at GSF peak pixel} \\
    a_i/N & \text{elsewhere}
\end{cases}
\end{equation}

\subsection{Bias Introduced by De-glare Operator}

The de-glare operator $\mathbf{S}$ introduced in Sec.~\ref{sec:sharpen} of is not a true inverse, and so will introduce bias in the recovered frames.  The relationship between the true, uncorrupted image $\mathbf{x}_t$ and its estimate $\mathbf{\hat{x}}_t^{(S)}$ can be obtained by plugging the expression

\begin{equation}
    \mathbf{\hat{y}}_t = \mathbf{A}\mathbf{x}_t = (1-\alpha)\mathbf{x}_t + \alpha \mathbf{B}\mathbf{x}_t .
\end{equation}

into Eq.~\ref{eq:deglarematrix}.  When we do so, we obtain the expression

\begin{equation}
    \label{eq:deglare_bias} 
    \mathbf{\hat{x}}_t^{(S)} = \mathbf{x}_t - \alpha^2 \left(\mathbb{I}-\mathbf{B}\right)^2 \mathbf{x}_t ,
\end{equation}

which consists of the true image and an additive bias term.  Inspection of Eq.~\ref{eq:deglare_bias} shows that the bias scales as a function of $\alpha^2$ and $\left(\mathbb{I}-\mathbf{B}\right)^2$.  Thus, the bias will remain small when the outscatter fraction $\alpha$ is small.  The bias is also strictly negative (assuming $\mathbf{x}_t$ is strictly positive), and takes the form of $\left(\mathbb{I}-\mathbf{B}\right)^2 \mathbf{x}_t$.



When the convolutional form of the de-glare operator (Eq.~\ref{eq:deglare_operator}) is used, additional bias will be introduced.  The severity of this bias will be low if the GSF is well-approximated as shift-invariant, and should increase with the degree to which that assumption is violated.

\label{sec:suppl_Learning_Approach}
\section{Baseline Implmentation Details}
Because we only have one baseline to compare to it is worth going into greater detail about the implementation and how we reconstructed their method. Following \cite{deep_learning_bloom_correction}, we implement a learning-based method that predicts per-pixel glare likelihoods. Let $H \in \mathbb{R}^{H \times W \times T}$ denote the input histogram volume (we use $T{=}672$ time bins). We extract a depth-index map $\hat{t} \in \{0,\dots,T{-}1\}^{H \times W}$ and an intensity map $I \in \mathbb{R}^{H \times W}$ via
\begin{align}
\hat{t}_{ij} &= \operatorname*{arg\,max}_{k} \; H_{ij[k]},\\
I_{ij} &= H_{ij[\hat{t}_{ij}]}.
\end{align}

\paragraph{GSF fitting.}
Given a transient histogram volume $H \in \mathbb{R}^{H \times W \times T}$ containing real-world glare artifacts, we first project it to a per-pixel depth map $D \in \mathbb{R}^{H \times W}$ and intensity map $I \in \mathbb{R}^{H \times W}$. In contrast to \cite{deep_learning_bloom_correction}, which operates directly on linear intensities $I$, we work in the log$_2$-intensity domain, since the intensity contrast between retroreflectors and background spans several orders of magnitude, making a direct exponential fit in linear space unstable. For each row containing an “aggressor’’ region (retroreflector pixels), we collect tuples $(\Delta x, \log r, w)$, where $\Delta x$ is the lateral distance from the aggressor edge, $w$ is the aggressor width for that row, and
\begin{align}
\log r = \log_2 \!\left(\frac{I_{\text{pixel}}}{I_{\text{peak}}}\right)
\end{align}
is the log-intensity ratio relative to the aggressor peak. Aggressor pixels themselves are excluded via a log-intensity threshold.

We normalize the distance by the aggressor width, $x = \Delta x / w$, aggregate all samples across rows, and fit a low-order polynomial $p(x)$ in the natural-log domain of the intensity ratio by solving
\begin{align}
\min_{p} \sum_i \bigl( p(x_i) - \ln r_i \bigr)^2,
\end{align}
where $\ln r_i = \log r_i \cdot \ln 2$. This yields a smooth decay function $\hat{r}(x) = \exp\!\bigl(p(x)\bigr)$ defined over $x \in [0,1]$.

\paragraph{Synthetic data generation.}

During synthesis, the glare intensity along each row is regenerated by evaluating the learned decay $\hat{r}(x)$ (with $x$ denoting the normalized lateral distance) and scaling the saturated retro intensity accordingly, while keeping the depth constant inside the glare band:
\begin{align}
    I_{\text{glare}}(x) &= I_{\text{retro}} \cdot \hat{r}(x), \\
    D_{\text{glare}}(x) &= D_{\text{retro}}.
\end{align}
This procedure produces synthetic depth and intensity maps, $D_{\text{synthetic}} \in \mathbb{R}^{H \times W}$ and $I_{\text{synthetic}} \in \mathbb{R}^{H \times W}$, that contain the synthetic retroreflector and its glare artifacts.

\paragraph{Network and training.}
We adopt a lightweight SqueezeSegV2 \cite{SqueezeSegV2} backbone that takes the two-channel tensor $[D_{\text{synthetic}} \,\|\, I_{\text{synthetic}}] \in \mathbb{R}^{H \times W \times 2}$ as input and outputs a binary glare likelihood map. The network is trained with class-balanced BCE-with-logits loss using Adam and a cosine-annealed learning rate schedule. We employ only horizontal/vertical flips as data augmentation. The model is trained for 50 epochs with a batch size of 4 on a dataset of 300 synthetic samples.

\paragraph{Inference and histogram-domain suppression.}
Since \cite{deep_learning_bloom_correction} outputs a glare-probability map $P \in [0,1]^{H \times W}$, we extend this baseline to the transient-histogram domain to enable a fair comparison with our method, which requires restoring the original depth and intensity maps from glare-corrupted data. Specifically, for any pixel whose confidence exceeds a fixed threshold ($P_{ij} \ge \tau$, with $\tau = 0.5$), we suppress the corresponding histogram bins within a full-width-at-half-maximum (FWHM) window centered at the estimated time bin $\hat{t}_{ij}$. Throughout our experiments, we set $\mathrm{FWHM} = 10$ time bins, matching the temporal FWHM of the emitted laser pulse in our LiDAR system and assuming it remains unchanged after reflection. Formally,
\begin{align}
H'_{ij[k]} =
\begin{cases}
0, & \text{if } \bigl|k - \hat{t}_{ij}\bigr| \le \frac{\mathrm{FWHM}}{2} \text{ and } P_{ij} \ge \tau, \\[3pt]
H_{ij[k]}, & \text{otherwise.}
\end{cases}
\end{align}
The resulting histogram $H'$ is thus ``glare-cleaned,'' in the sense that glare contributions are nulled while the rest of the temporal response remains intact.

\paragraph{Comparison to Our Method}
One major downside to this method is, as discussed in the main paper, that it operates in the depth map space which may be too late. Signals masked by glare may have already been removed during the postprocessing required to transform histograms to depth maps. Another downside is that this method requires retroreflective elements to be geometries that can be easily modeled parametrically (e.g. octagons, rectangles, circles). Objects with more complex geometry such as the cones and drums we tested underperformed. Finally, this method assumes a single retroreflector in the scene and does not adapt well to scenes with multiple retroreflectors, unlike our approach which is scene independent.

\end{document}